\documentclass[10pt,conference, final, 14paper, oneside, twocolumn]{IEEEtran}
\usepackage{graphicx}

\title{Knowledge driven Offline to Online Script Conversion}
\author{Sunil Kopparapu, Devanuj, Akhilesh Srivastava, P.V.S. Rao\\
Cognitive Systems Research Laboratory,\\
Tata Consultancy Services Limited,  Mumbai.}

\begin{document}
\maketitle

\begin{abstract}

The problem of offline to online script conversion is a challenging and 
an ill-posed problem. The interest in
offline to online conversion exists because there are a plethora of robust
algorithms in online script literature which can not be used on offline
scripts.  In this paper, we propose a method, based on heuristics, to extract 
online script information from offline bitmap image. We show the performance of
the proposed method on a real sample signature offline image, whose online
information is known.

\end{abstract}

\section{Background}

Offline script recognition by computers is well-addressed in literature and
still attracts attention from several researchers around the globe \cite{16,17}.
There is rich literature in both online and offline script research. 
The main difference
between online\footnote{Online signature is represented as a set of (x, y)
coordinates}
and offline\footnote{bitmap image in one of the several formats (.jpeg, .gif,
.pgm, .png, .bmp)}
signatures lies in the fact that online
signature captures the manner in which the signature was written while the
offline signature has no information on how the signature was generated or
written. 
In this paper, we give a brief survey of the offline script
literature. Recent surveys by Steinherz et. al \cite{17} for offline script and 
Lorigo and Govindraju \cite{16} specifically  for Arabic offline script show the interest
offline script recognition research carries.
Rigoli \cite{1} et al, compare the use of Hidden
Markov models (HMMs) for both on-line and off-line signature verification.
While HMM is a good model to capture second order statistics \cite{1}; the
use of HMM for signature verification is questionable. In practice for a
given person we can have limited number of signature samples (at maximum
three or four) to model the signature through its statistics. Sabourin and
Drouhard \cite{3} proposed the use of artificial neural networks (ANN) to
model signatures. Neural networks, like, HMMs need a large number of
samples to model the signature. Coetzer et. al \cite{5} use discrete radon
transform as the parameters to model the signature as HMMs. Leung et al
\cite{6} track features and the position of the stroke to verify
signatures. Peter et al \cite{10} use wavelet parameters to verify
signatures. where Qi et. al \cite{13} set up the problem of signature
verification in a multiresolution framework. They process the signature at
different resolution and then use the output at different resolutions to
verify signatures. Xiao et al \cite{14} use a modified Bayesian network
approach to verify signatures. While the approach suggested in each of
these references is based either on offline or on online signatures, there
have been approaches suggested \cite{15} which tend to take multi-modal
cues to verify signatures\footnote{multi-modal approach would require that the
same signature be available from different sources simultaneously}.

In this paper, we propose a method for deriving online information 
from  offline script. 
%This enables one to apply robust online signature
%verification procedures and algorithms available in literature to verify
%offline signatures \cite{1} . 
To the best of our knowledge, there is no
reported work in literature that deals with procedures to convert an offline
script into an online script.
 The closest work is
the work reported by Zimmer and Ling \cite{15} . They propose a hybrid
handwritten signature verification system where the on-line reference data
acquired through a digitizing tablet serves as the basis for the
segmentation process of the corresponding scanned off-line data. Local
foci of attention over the image are determined through a self-adjustable
learning process in order to pinpoint the feature extraction process. Both
local and global primitives are processed and the decision about the
authenticity of the specimen is defined through similarity measurements.

In Section \ref{sec:problem_formulation} we formulate the problem of deriving
online information from offline script, in Section \ref{sec:off2online} we
give details of the procedure adopted to derive online information followed by
experimental results in Section \ref{sec:experimental_results} and conclusions in
Section \ref{sec:conclusions}.

\section{Problem Formulation}
\label{sec:problem_formulation}

The problem that we address in this paper is one of extracting online
information from an offline script in the form of a bitmap image.
Essentially, we
need to derive the way the script was written by looking at the final
shape of the script. Clearly, this is an ill-posed problem\footnote{in the absence of the
actual online signature information there are several ways in which the
offline signature could have been generated}.

 \begin{figure}[h]
\includegraphics [width=0.5\textwidth] {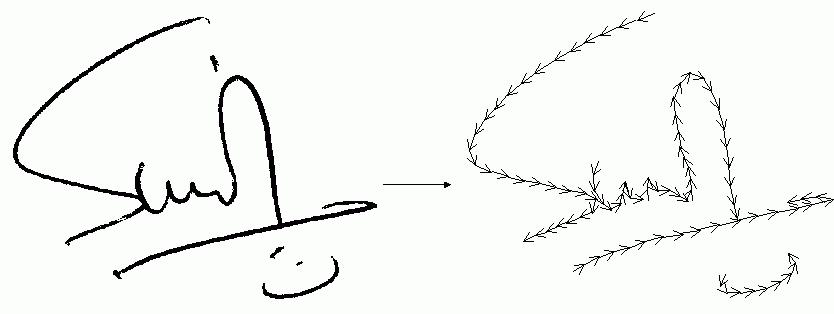} 
\caption{Offline signature (bitmap image) and the corresponding
online signature. Observe that the online signature has information on how the
signature was written (shown by arrow direction).}
\label{fig_1}
\end{figure}

 Observe that, 
 %\begin{enumerate}
 %\item shown only the offline bitmap image and
 shown only the bitmap image and
 %\item in the absence of knowledge about English script, there are several
 in the absence of knowledge about English script, there are several
ways in which the signature could have been written. While there is no
such ambiguity in determining how the signature was written when 
online information is available. The problem of offline to online signature conversion is
to identify the way the signature was written from the bitmap
image\footnote{We assume that one is aware of the way script is written;
essentially the language in which the signature is written.}.
% \end{enumerate}
Fig. \ref{fig_1} shows an offline signature and its equivalent online
information. The direction of  the arrow shows the way the online signature was
written. It also captures the number of strokes (four in this case) 
in the signature (pen-lifts). How ever what is not depicted is the order in
which the four segments were written. In this paper we do not address the
problem of identifying the order in which the segments were written.

For script belonging to the same set\footnote{Same language, English
for instance}, it is possible to traverse the signature using rules based
on heuristics. These derived rules would primarily be based on the
knowledge of how people write that script (left to right, top to bottom,
etc) in that particular language.  A large portion of such rules would be
based primarily on the language in which the signature is written in
addition to heuristics.

\section{Offline to Online Conversion}
\label{sec:off2online}

The bitmap offline signature image consists of the actual
signature (the written part) and the background (the paper on which the
signature was signed). The primary idea of our methodology is to
intelligently traverse the written part in the bitmap image signature. We 
 assume the script/signature to form a path/road and the traversal scheme being a
truck driver who is trying to stay on the path. The driver of the vehicle
steers the truck along the signature path so as to stay on the path all the time.

 \begin{figure}[h]
 \centerline{\includegraphics [width=0.5\textwidth] {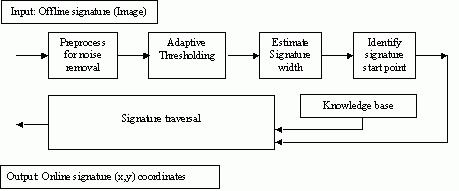} }
 \caption{Block diagram of the offline to online conversion process }
 \label{fig_2}
 \end{figure}

Block diagram of the complete offline to online conversion process is
shown in Fig. \ref{fig_2}. The offline signature is obtained
by scanning a signature on a paper  with the help of a scanner. The
original offline bitmap signature images are normally $8$ bit which
results in any of the pixel in the image taking a value between $0$ and
$255$, namely, $256$ ($2^8$) gray pixel intensity levels. The first step
in the offline to online signature conversion is to preprocess the bitmap
image using some basic image processing to remove any noise that might
have cropped up in the scanning process. We use a $5 \times 5$ 
median filter to remove noise. 
This gray level image is binarized using dynamic thresholding method. The
threshold is determined by observing that the histogram plot of the gray level
image would largely be a two hump plot. 
 We take the two
highest peaks\footnote{one peak is due to the foreground and the other
peak is due to the background} in the pixel- intensity histogram of the
bitmap and record the corresponding intensities. The binarizing threshold
is set at the intensity that lies at the middle of the two intensities.
The actual signature traversal\footnote{which enables conversion to online
signature} is carried out on the binary image.

%The signature (actual sign) mark can be
%termed as the foreground while the vast unmarked area (the paper on which
%the signature was marked) can be looked upon as the background. Obviously,
%the number of background pixels far exceeds those of the foreground. This
%information is used to binarize the signature bitmap.

The binarised image is then traversed using the truck driver steering his
truck on the signature. 
The traversal process that we have adopted is influenced by the way a
truck driver steers the truck when driving the truck on a road. 
The start of the road or the signature is determined by scanning the bitmap
image from top to bottom and from left to right (probably one would adopt
another strategy if one were to look at script in a different language). The
first road pixel becomes the start point of the truck.
Now the strategy adopted by truck driver is to
steering the truck such that the truck stays on and in the middle of the
road.  
 \begin{figure}[h]
 \includegraphics [width=0.5\textwidth] {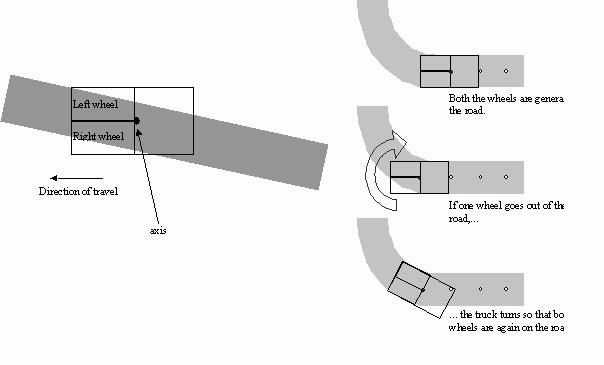}
 \caption{Construction of a virtual truck and the virtual truck
steering itself to stay in the middle of the road when it is tending to
get off the road. }
 \label{fig_3}
 \end{figure}
We assume the written part of the
 signature to be the road to be traversed. We construct a virtual
truck\footnote{The size of the virtual truck is different for different
signature traversal. The size of the truck is a function of the width of
the signature.} with two wheels, which sense if the truck is going off the
road (see Fig. \ref{fig_3}) by determining the ration of the road pixels under
each of the wheels. When the truck is tending to get off the
road (namely the number of road pixels under the left and right truck are not
approximately same) 
it steers itself so that the number of pixels under both the left and the right
wheel are same and hence stays on the middle of the
road.  This allows the virtual truck to traverse the signature.
%The wheels help in examining and revising the truck's orientation
%vis-à-vis the road simply by comparing the number of road-pixels under
%each of them.
%This approach, no doubt, has its own set of problems, for example, the no
%of pixels under the wheels becoming so low that comparisons cannot be done
%effectively. 
The places of intersection in the signature, where
two paths cross each other, both the paths become available for the truck to
steer itself. In such situations, we direct the truck to  
proceed in the direction in which it has been moving.
%we are
%traversing  to the path being travelled can make the
%turn turn towards itself.
 
\begin{figure}[h]
 \centerline{\includegraphics [width=0.5\textwidth] {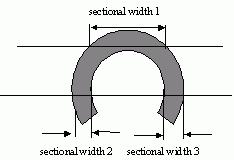}}
 \caption{ Determining the average width of the signature. The average
width of the signature is used to determine the size of the conceptual
truck used to traverse the signature. }
 \label{fig_4}
 \end{figure}

In order that the truck is able to traverse a signature accurately, the
two wheels of the truck should be as close to the corresponding two
edges of the signature. The thickness of different signatures is not uniform 
because of different writing material used. 
This requires that the size of the truck be a function of the
thickness of the signature. The size of the truck is dependent 
on the average width of the signature.
The average signature width is calculated (see Fig.  \ref{fig_4}) as the normalized average of highest
three sectional widths with respect to the $x$-axis. Suppose there are $n$
sectional widths estimated in a signature image. and let $X_i$ be the
sectional width of the $i^{th}$ section. Without loss of generality we can
assume that $X_i$ is arranged in the increasing order of widths for
increasing $i$. The average width of the signature is calculated as
\begin{equation}
\frac{\sum_{i=n-3}^{n} X_i \times i}{\sum_{i=n-3}^{n} X_i}
\label{eq:width}
\end{equation}
{\bf Note:} The sectional width can be looked upon as the distance of an
uninterrupted stretch
 of signature (foreground pixels) if one travels along a line parallel to
x-axis. For example, the sectional width is the number of pixels between
the first point where the foreground changes to background and the next
immediate instance when the background changes to foreground (see Fig.
\ref{fig_4}).

\section{Experimental Results}
\label{sec:experimental_results}

\begin{figure}[h]
 \includegraphics [width=0.5\textwidth] {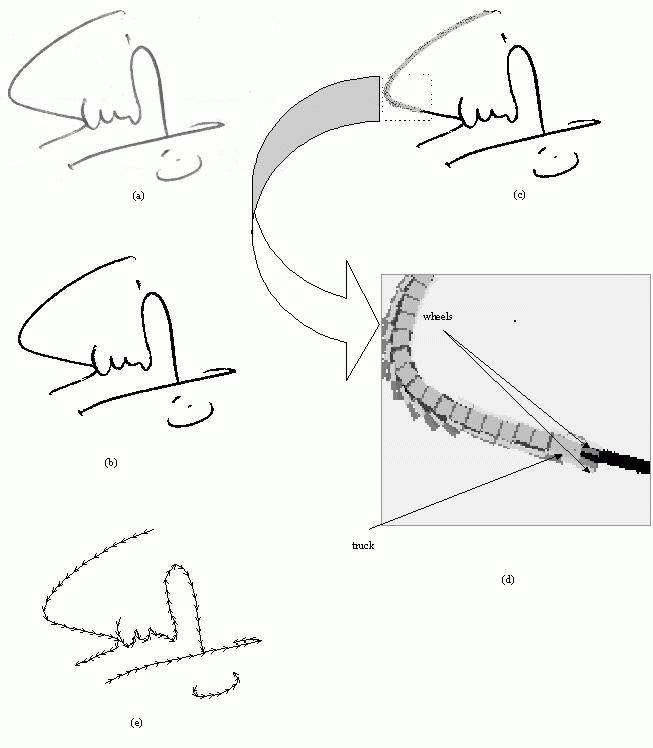}
 \caption{ Process of extracting online information from the offline
signature bitmap for a sample signature. (a) Original bitmap image (256
gray levels), (b) binarized image (2 gray levels), (c) initial traversal,
(d) expanded view of initial traversal, (e) fully traversed image. }
 \label{fig_5}
 \end{figure}

We tested the robustness of the developed scheme by capturing online
signature from CrossPad\footnote{IBM manufactured online signature
capturing device}. The CrossPad in addition to capturing the online
signature in the form of $(x,y)$ coordinates also supports the capture of
the same image in the form of a bitmap image. We used the bitmap signature
image as input to our offline to online conversion system and cross check
the generated online data with the online $(x, y)$ information given by the
CrossPad. Fig. \ref{fig_5} shows one of the offline
signatures
%\footnote{of one of the authors} 
on which we tested our
scheme\footnote{We have tested the scheme on several signature images. We will
give more experimental results on different offline images in the final paper}

Notice that there are three major steps involved in the process of offline
to online conversion. Initially, the bitmap image is binarized (bi-gray
value image), the truck dimension are estimated from the average width of
the signature (using \ref{eq:width})
and then using heuristics (Section \ref{sec:off2online}) 
the truck is made to traverse the
signature bitmap. The whole process is captured in Fig. \ref{fig_4} for
a sample signature image. We compare the obtained online information with
the actual online data in terms of the number of strokes (pen lifts) and
the direction of traversal. The trace of each segment was same as that captured
in the online information from CrossPad. Nevertheless, as discussed earlier the
segments were not traversed in the same order in which the signature was
written. The segment three (the
long line in the signature)  was traced as segment two. More work is being
carried out in this regard as part of our ongoing research.

%\begin{figure}[h]
%\includegraphics [width=0.5\textwidth] {figs/fig_06.jpg}
%\caption{A typical crossover point}
%\label{fig_6}
%\end{figure}
 
%\subsection{Issues}
%
%One of the major issues is in identifying the start point of the
%signature. There is no robust way of determining this. We search for the
%start of the signature by traversing from right to left and top to bottom
%with the that one end of the chosen start point be adjacent to the
%background. This method seems to work well most of the time.  

%The other is
%the crossover point (Fig \ref{fig_6}). Initially, we identify the crossover points
%in the signature and then we adopt the strategy of allowing the truck to
%follow the direction in which it entered the crossover point. This
%strategy seems to work very well. The strategy adopted in based on the
%observation that human rarely change the direction of the strokes
%abruptly.

\section{Conclusion}
\label{sec:conclusions}

In this paper, we have proposed a heuristics based technique to convert 
an offline
signature to an online signature by extracting the trace information of
the signature from the offline bitmap image. The motivation for offline to
online conversion comes from the fact that algorithms for online signature
verification are more robust than the offline signature verification
systems. This is true primarily because of the extra trace information
that is absent in the offline bitmap images. The knowledge driven offline
to online conversion algorithm described in this paper is able to robustly
trace the signature in the bitmap image to produce an online image. It is
proposed that the derived online information from the offline image be
used to perform computer signature verification using several of the
online signature verification algorithms proposed in literature.

%\section{References}

\end{document}